# Efficient Chambolle-Pock based algorithms for Convoltional sparse representation

Yi Liu*, Junjing Li, Yang Chen, *Senior Member, IEEE*, Haowei Tang, Pengcheng Zhang, Tianling Lyu*, Zhiguo Gui

*Abstract*—Recently convolutional sparse representation (CSR), as a sparse representation technique, has attracted increasing attention in the field of image processing, due to its good characteristic of translate-invariance. The content of CSR usually consists of convolutional sparse coding (CSC) and convolutional dictionary learning (CDL), and many studies focus on how to solve the corresponding optimization problems. At present, the most efficient optimization scheme for CSC is based on the alternating direction method of multipliers (ADMM). However, the ADMM-based approach involves a penalty parameter that needs to be carefully selected, and improper parameter selection may result in either no convergence or very slow convergence. In this paper, a novel fast and efficient CSC method using Chambolle-Pock(CP) framework is proposed, which does not require extra manual selection parameters in solving processing, and has faster convergence speed. Furthermore, we propose an anisotropic total variation penalty of the coefficient maps for CSC and apply the CP algorithm to solve it. In addition, we also apply the CP framework to solve the corresponding CDL problem. Experiments show that for noise-free image the proposed CSC algorithms can achieve rival results of the latest ADMM-based approach, while outperforms in removing noise from Gaussian noise pollution image.

## I. Introduction

CONVOLUTIONAL sparse representation (CSR) is a recently widely used representation in image processing, it is widely thought to originate from the deconvolutional networks of Zeiler [1], and is an convolutional form of traditional sparse representation (based on image patch)[2]. Different from the multiply operation in traditional sparse representation, CSR models an image through the convolution sum of a set of linear filters and corresponding coefficient maps [3]. In this case, an image $\mathbf{s}$ with size of $H \times W$ can be approximated by $\mathbf{s} \approx \sum_{m=1}^{M} \mathbf{d}_m * \mathbf{x}_m$, where $*$ indicates the convolutional operation, $\{\mathbf{d}_m\} \in \mathbb{R}^{r \times r}$ is a set of $M$ linear filters and each filter is size of $r \times r$, $\mathbf{x}_m \in \mathbb{R}^{H \times W}$ is the corresponding coefficient map for the $m$-th filter $\mathbf{d}_m$, and a set of coefficient maps $\{\mathbf{x}_m\}_{m=1}^{M}$ is formed with $M$ filters. Since $\mathbf{s}$ is represented in convolution form and is an entire image instead of an image patch (which is reconstructed in traditional sparse representation), CSR is a single-valued translation-invariant sparse representation, thus overcomes the over-smoothing and artifacts caused by average arrangement of overlapping image patches in traditional sparse representation [4]. In view of these advantages, CSR has been widely used in image processing, such as image fusion [5-7], image reconstruction [8, 9], and image denoising [10-12], etc.

Obviously, the content of CSR includes two parts: (1) Convolutional sparse coding (CSC), incorporates a minimization model to find the optimal sparse coefficient maps $\{\mathbf{x}_m\}_{m=1}^{M}$ to represents the entire image, with a given set of filters $\{\mathbf{d}_m\}_{m=1}^{M}$ (2) Convolutional dictionary learning (CDL), namely how to get the optimal convolutional dictionary (a set of filters $\{\mathbf{d}_m\}_{m=1}^{M}$) given the succeeding coefficient maps. The unconstrained CSC problem is usually posed with a $\ell_1$ norm of coefficient maps and typically given by the following form

$$\arg\min_{\{\mathbf{x}_m\}} \frac{1}{2} \left\| \sum_{m}^{M} \mathbf{d}_m * \mathbf{x}_m - \mathbf{s} \right\|_2^2 + \lambda \sum_{m}^{M} \|\mathbf{x}_m\|_1, \quad (1)$$

where $\lambda$ is a parameter that constrains the sparsity of coefficient maps, $\mathbf{s}$ and $\mathbf{x}_m$ are considered to be vectored into $N$ dimensions ($N = H \times W$). To solve the minimization problem in (1), Zeiler introduced an auxiliary variable and split (1) into two subproblems, then the strategy with alternating minimization of two subproblems are utilized. However, this method has a heavy computational burden [1]. To solve this problem, Chalasani *et al* [13] utilized the fast iterative shrinkage thresholding algorithm (FISTA) [14]

This work was supported in part by the Natural Science Foundation of Shanxi Province under Grant 202303021211148 and 202103021224204, the National Key Research and Devel opment Program of China under Grant 2022YFE0116700, and the National Natural Science Foundation of China under Grant T2225025 (Corresponding author: Yi Liu).

Yi Liu, Junjing Li, Haowei Tang, Pengcheng Zhang, Zhiguo Gui are with the State Key Laboratory of Dynamic Testing Technology, North University of China, Taiyuan, China, and also with School of Information and Communication Engineering, North University of China, Taiyuan, China(liuyi@nuc.edu.cn,ljj06062000@163.com, tanghaowei77@163.com, zhangpc198456@163.com, guizhiguo@nuc.eu.cn).

Yang Chen is with the Jiangsu Provincial Joint International Research Laboratory of Medical Information Processing, School of Computer Science and Engineering, Southeast University, Key Laboratory of New Generation Artificial Intelligence Technology and Its Interdisciplinary Applications (Southeast University), Ministry of Education, China (e-mail:chenyang.list@seu.edu.cn).

Tianling Lyu is with Academy for Advanced Interdisciplinary Science and Technology, Zhejiang University of Technology, Hangzhou, Zhejiang, China (e-mail: lyutianling@zjut.edu.cn).

as an alternative solving technique. Almost simultaneously Bristow *et al* [15] proposed to use the alternating direction method of multipliers (ADMM) framework [16] for CSC, and obtained a fast convergence process and speeded up calculation. Since then a variety of ADMM-based approaches have been proposed for CSC algorithms[17-22], and especially in Bristow's subsequent work [20], the proposed ADMM algorithm was proved to be much faster than FISTA method in [13]. At present the most efficient CSC algorithms are all most based on ADMM.

ADMM splits the CSC problem into two sub-problems by introducing an auxiliary variable. One sub-problem is a sparse approximation problem of the auxiliary variable, and can be easily solved using a shrinkage thresholding operator. The other sub-problem that involves a convolutional $\ell_2$ data fidelity term and a penalty of coefficient maps is more troublesome, an efficient solution is to transform the convolutional system to a linear system via convolution theorem, thus this sub-problem becomes solving a very large linear system. The major difference between these ADMM-based methods consists in solutions for the linear system in DFT (Discrete Fourier Transform) domain. These solutions include conjugate gradient, gaussian elimination [15], Sherman-Morrison [3,18,19], and the recent state-of-the-art method in [22] to further reduce the computational expense. The former two solutions need to be implemented iteratively, while the latter two solutions are non-iterative and have less complexity, especially the method in [22].

The CDL problem usually consists of two steps, a CSC step given the current dictionary and a dictionary update step given the resulting sparse coefficient maps. These two steps are alternately performed util a stopping criteria is met, thus the CDL problem is more complex than CSC. The CDL problem is usually formed as

$$\arg\min_{\{\mathbf{x}_{m,v}\},\{\mathbf{d}_m\}} \sum_{v=1}^{V}\left(\frac{1}{2}\left\|\sum_{m=1}^{M}\mathbf{d}_m * \mathbf{x}_{m,v} - \mathbf{s}_v\right\|_2^2 + \lambda \sum_{m=1}^{M}\|\mathbf{x}_{m,v}\|_1\right) \quad (2)$$

$$s.t. \quad \mathbf{d}_m \in \Omega \quad \forall m,$$

where $\mathbf{s}_v$ is the $v$-th image from a training set of $V$ images, $\mathbf{x}_{m,v}$ refers to the coefficient maps of $\mathbf{d}_m$ for the $v$-th image. $\Omega = \{\mathbf{d}_m \mid \|\mathbf{d}_m\|_2 = 1, m = 1, \cdots M\}$ is the constraint on the norms of filters to avoid the scaling ambiguity between coefficients and filters. The common approach to solve Eq.(2) is to do alternate minimization between CSC step and dictionary update step. In the CSC processing, for multiple images, the minimization problem is the multiple measurement vector (MMV) version of Eq.(1), while the minimization with respect to dictionary filters $\{\mathbf{d}_m\}$ can be described as follows

$$\arg\min_{\{\mathbf{d}_m\}} \frac{1}{2}\sum_{v=1}^{V}\left\|\sum_{m=1}^{M}\mathbf{d}_m * \mathbf{x}_{m,v} - \mathbf{s}_v\right\|_2^2 \quad s.t. \quad \mathbf{d}_m \in \Omega \quad \forall m. \quad (3)$$

The earlier algorithms for CDL mainly focus on convolutional extensions of the traditional dictionary update using K-SVD methods [23-25], block coordinate method [26], and gradient descent based methods [27-29]. These algorithms are less effective and time-consuming, since they always consist of outer alternation between a CSC step and a dictionary update step, both of which require inner update iterations. The most efficient solvers for Eq.(3) are FISTA [30, 31, 36] and ADMM [17, 19, 20, 32-38], and the dictionary update step is always performed in DFT domain. In the context of ADMM, as in the case of CSC, different approaches can be used to solve the linear system related to dictionary filters, theses algorithms includes conjugate gradient, iterated Sherman-Morrison [19], and spatial tiling [39].

Whether CSC or CDL, ADMM framework is an effective solution, nevertheless, the ADMM algorithm involves a penalty parameter $\rho$ and the performance of ADMM is easily affected by the penalty parameter, an improper parameter selection may result in either no convergence or very slow convergence. Chambolle-Pock(CP) [40] is a first-order primal-dual algorithm that can solve a convex problem simultaneously with its dual problem, and has convergence guarantee [41]. In this paper, we abandon the popular ADMM framework but propose a novel efficient CSC method by using the CP framework. In the CP framework, the CSC solving processing only involve a single-layer loop by using the proximal mapping, and can improve the computational efficiency for minimizing Eq.(1). Besides, In addition, other than the number of iterations, no other hyperparameters need to be manually selected in the CP framework. In addition, to enhance the performance of convolutional sparse representation, we propose an anisotropic TV penalty of the coefficient maps and expand the corresponding CP-based coding (described in the Section III.B). For CDL, we propose to apply the CP framework to update the dictionary filters in the CDL process. The main contribution of this paper is to introduce CP framework to solve CSC and CDL problems for the first time. To our best knowledge, there are no previous papers utilizing CP algorithm for CSC and CDL, and this study is the first exhibition.

The remainder of the paper is organized as follows. Section II briefly reviews CP algorithm. In Section III, The proposed CSC methods and CDL method are elaborated. Experimental evaluation and comparisons in terms of reconstruction accuracy and computation complexity are presented in Section IV. In Section V, we conclude this paper.

## II. CP Algorithm

The CP framework is often used to solve primal minimization problem in the following general form [41]

$$\min_{x \in X}\{F(Kx) + G(x)\}, \quad (4)$$

where $x$ is a finite real vector in space $X$, $K$ is a linear map, functions $G$ and $F$ are convex, but not required smooth. The dual maximization problem corresponding to Eq.(4) is as follows

$$\max_{y \in Y}\{-F^*(y) - G^*(-K^T y)\}, \quad (5)$$

where $y$ is a finite real vectors in $Y$ and $y = Kx$. $F^*$ and $G^*$ are convex conjugation functions of $F$ and $G$, respectively. Symbol '$T$' means transpose operation. Given a convex function $H(z)$, the definition of conjugate convex function $H^*(z)$ is following

$$H^*(z) = \max_c \{\langle z, c \rangle - H(c)\}, \quad (6)$$

where $\langle z, c \rangle$ indicates an inner product.

The primal-dual problem described above is formally connected in the following saddle-point optimization problem in Eq.(7)

$$\min_x \max_y \{\langle Kx, y \rangle + G(x) - F^*(y)\}. \quad (7)$$

The saddle-point optimization problem is solved by performing the maximization over dual variable $y$ and then minimization over primal variable $x$. The splitting maximization and minimization problems are described in Eq. (8) and Eq. (9).

$$\max_y \{\langle Kx, y \rangle - F^*(y)\}, \quad (8)$$

$$\min_x \{\langle Kx, y \rangle + G(x)\}, \quad (9)$$

The idea of CP algorithm is to apply proximal gradient method [42] to two variables. The proximal operator is often marked as $prox(\cdot)$ and is obtained by the following minimization

$$prox_\sigma[H](z) = \arg\min_{z'} \left\{ H(z') + \frac{1}{2\sigma} \|z - z'\|_2^2 \right\}. \quad (10)$$

By using the proximal gradient upcent for Eq.(8) and descent for Eq.(9), respectively, we have the iterative update formulas in Eq. (11) and Eq.(12), and the CP framework summarized in [42] is shown in **Algorithm 1**. As we see, the CP algorithm is simple and allows both primal variable and dual variable to be updated simultaneously.

$$y_{n+1} = \arg\max_y \left\{ -F^*(y) + \langle Kx_n, y - y_n \rangle - \frac{1}{2\sigma}\|y_n - y\|_2^2 \right\} \quad (11)$$
$$= prox_\sigma[F^*](y_n + \sigma K x_n),$$

$$x_{n+1} = \arg\min_x \left\{ G(x) + (y_{n+1})^T K(x - x_n) + \frac{1}{2\tau}\|x_n - x\|_2^2 \right\} \quad (12)$$
$$= prox_\tau[G](x_n - \tau K^T y_{n+1}).$$

In **Algorithm 1**, the non-negative parameter $L$ is obtained by the largest singular value of matrix $K$, and the calculation of $L$ is detailed described in [41]. $\tau$ and $\sigma$ are set to $1/L$ in practice.

The CP algorithm has been proved to solve many regularization problems such as $\ell_1$-regularization, TV-regularization, TGV-regularization, and so on, thus it has been widely used method in various fields, including image processing[43, 44] computer vision[45], and image reconstruction[46, 47].

---

**Algorithm 1**   Generic CP framework [41]
1:  $L = \|K\|_2; \tau = 1/L; \sigma = 1/L; \theta = 1$
2:  initialize $x_0$ and $y_0$ to zero values
3:  $\bar{x}_0 = x_0$
4:  for $n = 1$: iterN
5:      $y_{n+1} = prox_\sigma[F^*](y_n + \sigma K \bar{x}_n)$
6:      $x_{n+1} = prox_\tau[G](x_n - \tau K^T y_{n+1})$
7:      $\bar{x}_{n+1} = x_{n+1} + \theta(x_{n+1} - x_n)$
8:  end

---

## III. PROPOSED METHODS

In view of the successful application of CP for convex optimization problems, it is a natural choice for CSC. The iterative update formulas of CP algorithm for CSC and CDL problems are derived in this section.

### A. CP Algorithm for Unconstrained CSC

Similar to the previous study in [18,36], we define linear operators $D_m$ such that $D_m \mathbf{x}_m = \mathbf{d}_m * \mathbf{x}_m$, and

$$D = (D_0 \quad D_1 \quad \ldots) \quad \mathbf{x} = \begin{pmatrix} \mathbf{x}_0 \\ \mathbf{x}_1 \\ \vdots \end{pmatrix}, \quad (13)$$

then we can rewrite the unconstrained CSC problem in Eq.(1) as follows

$$\arg\min_{\mathbf{x}} \frac{1}{2}\|D\mathbf{x} - \mathbf{s}\|_2^2 + \lambda \|\mathbf{x}\|_1. \quad (14)$$

To derive the CP algorithm for problem Eq.(14), we do the following transformations by associating Eq. (14) with Eq. (4), namely let

$$K = D, \quad \mathbf{y} = D\mathbf{x},$$

$$F(\mathbf{y}) = \frac{1}{2}\|\mathbf{y} - \mathbf{s}\|_2^2, \quad G(\mathbf{x}) = \lambda \|\mathbf{x}\|_1 \quad (15)$$

Applying Eq.(6), we obtain the convex conjugates of $F$

$$F^*(\mathbf{p}) = \max_{\mathbf{p}'} \left\{ \langle \mathbf{p}, \mathbf{p}' \rangle - \frac{1}{2}\|\mathbf{p}' - \mathbf{s}\|_2^2 \right\} = \frac{1}{2}\|\mathbf{p}\|_2^2 + \langle \mathbf{p}, \mathbf{s} \rangle \quad (16)$$

To attain the CP algorithm instance for CSC problem, we derive the proximal mapping of $F^*(\mathbf{y})$ and $G(\mathbf{x})$ according to Eq.(10), and we can obtain:

$$prox_\sigma[F^*](\mathbf{y}) = \arg\min_{\mathbf{y}'} \left\{ \frac{1}{2}\|\mathbf{y}'\|_2^2 + \langle \mathbf{y}', \mathbf{s} \rangle + \frac{\|\mathbf{y} - \mathbf{y}'\|_2^2}{2\sigma} \right\} \quad (17)$$
$$= \frac{\mathbf{y} - \sigma \mathbf{s}}{1 + \sigma}$$

$$prox_\tau[G](\mathbf{x}) = \arg\min_{\mathbf{x}'} \left\{ \lambda \|\mathbf{x}'\|_1 + \frac{\|\mathbf{x} - \mathbf{x}'\|_2^2}{2\tau} \right\} \quad (18)$$
$$= sign(\mathbf{x}) \odot \max(|\mathbf{x}| - \tau \lambda, 0)$$

where $sign(\cdot)$ is the sign function and $\odot$ indicates Hadamard operation (element-wise multiplication). Although the proximal mapping results from a quadratic minimization, Eq. (17) and Eq. (18) both have closed-form solutions, thus it leads to a single-layer loop and is easy to update iteratively. By Substituting the arguments from the generic CP algorithm, we can attain the corresponding CP algorithm for the unconstrained CSC problem and summary it in **Algorithm** 2. For the sake of description, we name this proposed method "CSC-CP".

---
**Algorithm 2** CSC-CP algorithm
---
1: Setting $L=\|D\|_2; \tau=1/L; \sigma=1/L; \theta=1$
2: Initialize $\mathbf{x}_0, \mathbf{p}_0$ to zero values
3: $\overline{\mathbf{x}}_0 = \mathbf{x}_0$
4: for $n=1:\text{iter}N$
5: $\quad \mathbf{p}_{n+1} = prox_\sigma[F^*](\mathbf{p}_n + \sigma D\overline{\mathbf{x}}_n)$
$\quad\quad = (\mathbf{p}_n + \sigma(D\overline{\mathbf{x}}_n - \mathbf{s}))/(1+\sigma)$
6: $\quad \mathbf{x}_{n+1} = prox_\tau[G](\mathbf{x}_n - \tau D^T \mathbf{p}_{n+1})$
$\quad\quad = sign(\mathbf{x}_n - \tau D^T \mathbf{p}_{n+1})\max(|\mathbf{x}_n - \tau D^T \mathbf{p}_{n+1}| - \tau\lambda, 0)$
7: $\quad \overline{\mathbf{x}}_{n+1} = \mathbf{x}_{n+1} + \theta(\mathbf{x}_{n+1} - \mathbf{x}_n)$
8: end
---

Note that lines 5 and 6 in **Algorithm** 2 involve convolution and transpose convolution, to improve computing efficiency, we perform them in the frequency domain according to the convolution theorem. In detail, we exploit the FFT for convolution and transpose convolution, and then exploit the inverse FFT to obtain the temporary image $D\overline{\mathbf{x}}$ and coefficient maps $D^T\mathbf{y}$. It should be noted that the size of $\mathbf{d}_m$ is much smaller than that of $\mathbf{x}_m$ and $\mathbf{s}$, thus $\mathbf{d}_m$ needs to be zero-padded to the spatial dimensions of $\mathbf{x}_m$ and $\mathbf{s}$. In a single iteration, the dominated computational cost is $\mathcal{O}((M+1)N\log N)$ from FFTs, and the computational cost of proximal mapping in Eq.(18) is $\mathcal{O}(MN)$. It is worth noting that apart from the necessary FFT operation, there are only scalar multiplication and subtraction in the spatial domain.

*B. CP Algorihtm for CSC with Anisotropic TV Penalty*

To enhance the performance of convolutional sparse representations, we propose an anisotropic TV penalty of the coefficient maps to extend Eq.(1), and apply the CP algorithm to solve the extended CSC problem, called CSCATV-CP. The proposed CSC with anisotropic TV penalty is described as

$$\arg\min_{\{\mathbf{x}_m\}} \frac{1}{2}\left\|\sum_m^M \mathbf{d}_m * \mathbf{x}_m - \mathbf{s}\right\|_2^2 + \lambda\sum_m^M \|\mathbf{x}_m\|_1 + \sum_m^M \left(\beta_1 \|(|G_0\mathbf{x}_m|)\|_1 + \beta_2 \|(|G_1\mathbf{x}_m|)\|_1\right), \quad (19)$$

where $G_0$ and $G_1$ are discrete differential operators in horizontal and vertical directions, respectively, $\beta_1$ and $\beta_2$ are the corresponding weights for gradient penalties of the coefficient maps. By introducing block matrix notation, Eq.(19) can be rewritten as follow:

$$\arg\min_{\mathbf{x}} \frac{1}{2}\|D\mathbf{x}-\mathbf{s}\|_2^2 + \lambda\|\mathbf{x}\|_1 + \beta_1 \|(|\Phi_0\mathbf{x}|)\|_1 + \beta_2 \|(|\Phi_1\mathbf{x}|)\|_1, \quad (20)$$

where

$$\Phi_k = \begin{pmatrix} G_k & 0 & \cdots \\ 0 & G_k & \cdots \\ \vdots & \vdots & \ddots \end{pmatrix}. \quad (21)$$

To apply CP algorithm to Eq.(20), we carry out the following assignments:

$$F(\mathbf{y},\mathbf{v},\mathbf{z}) = F_1(\mathbf{y}) + F_2(\mathbf{v}) + F_3(\mathbf{z}), \quad (22)$$

$$F_1(\mathbf{y}) = \frac{1}{2}\|\mathbf{y}-\mathbf{s}\|_2^2, F_2(\mathbf{v}) = \beta_1\|(|\mathbf{v}|)\|_1,$$
$$F_3(\mathbf{z}) = \beta_2\|(|\mathbf{z}|)\|_1, G(\mathbf{x}) = \lambda\|\mathbf{x}\|_1, \quad (23)$$

$$\mathbf{y} = D\mathbf{x}, \mathbf{v} = \Phi_0\mathbf{x}, \mathbf{z} = \Phi_1\mathbf{x}, K = \begin{pmatrix} D \\ \Phi_0 \\ \Phi_1 \end{pmatrix}. \quad (24)$$

In order to get the convex conjugation of $F$, we also need to obtain $F_2^*, F_3^*$. By the definition in Eq.(6), we have

$$F_2^*(\mathbf{q}) = \max_{\mathbf{q}'}\{\langle\mathbf{q},\mathbf{q}'\rangle - \beta_1\|(|\mathbf{q}'|)\|_1\}$$
$$= \delta_{Box(\beta_1)}(|\mathbf{q}|) \quad (25)$$

and

$$F_3^*(\mathbf{r}) = \max_{\mathbf{r}'}\{\langle\mathbf{r},\mathbf{r}'\rangle - \beta_2\|(|\mathbf{r}'|)\|_1\}$$
$$= \delta_{Box(\beta_2)}(|\mathbf{r}|) \quad (26)$$

where $\delta_{Box(\beta)}$ is the indicator function defined as follows:

$$\delta_{Box(\beta)}(x) \equiv \begin{cases} 0 & \|x\|_\infty \leq \beta \\ \infty & \|x\|_\infty > \beta \end{cases}. \quad (27)$$

According to Eq.(10), the proximal mappings of anisotropic TV penalty can be obtained as

$$prox_\sigma[F_2^*](\mathbf{v}) = \frac{\beta_1\mathbf{v}}{\max(\beta_1\mathbf{1}_I, |\mathbf{v}|)}, \quad (29)$$

$$prox_\sigma[F_3^*](\mathbf{z}) \frac{\beta_2\mathbf{z}}{\max(\beta_2\mathbf{1}_I, |\mathbf{z}|)}, \quad (30)$$

where $\mathbf{1}_I$ is an image with all pixels set to 1. Since the proximal mapping of the data item has been derived previously, the final proximal mapping by putting them together is described as

$$prox_\sigma[F^*](\mathbf{y},\mathbf{v},\mathbf{z}) = \left(\frac{\mathbf{y}-\sigma\mathbf{s}}{1+\sigma}, \frac{\beta_1\mathbf{v}}{\max(\beta_1\mathbf{1}_I, |\mathbf{v}|)}, \frac{\beta_2\mathbf{z}}{\max(\beta_2\mathbf{1}_I, |\mathbf{z}|)}\right). \quad (31)$$

The proximal mapping of $G(\mathbf{x})$ is the same as that in Eq.(18), and the iterative pseudocode of the proposed

CSCATV-CP is exhibited in **Algorithm 3**.

---
**Algorithm 3**     CSCATV-CP algorithm
---
1:    $L = \|D, \Phi_0, \Phi_1\|_2; \tau = 1/L; \sigma = 1/L; \theta = 1$
2:    Initialize $\mathbf{x}_0$, $\mathbf{z}_0$ to zero values
3:    $\bar{\mathbf{x}}_0 = \mathbf{x}_0$
4:    for $n=1$: iter$N$
5:      $\mathbf{p}_{n+1} = prox_\sigma[F_1^*](\mathbf{p}_n + \sigma D\bar{\mathbf{x}}_n)$
       $= (\mathbf{p}_n + \sigma(D\bar{\mathbf{x}}_n - \mathbf{s}))/(1+\sigma)$
6:      $\mathbf{q}_{n+1} = prox_\sigma[F_2^*](\mathbf{q}_n + \sigma \Phi_0 \bar{\mathbf{x}}_n)$
       $= \dfrac{\beta_1(\mathbf{q}_n + \sigma \Phi_0 \bar{\mathbf{x}}_n)}{\max(\beta_1 \mathbf{1}_I, |\mathbf{q}_n + \sigma \Phi_0 \bar{\mathbf{x}}_n|)}$
7:      $\mathbf{r}_{n+1} = prox_\sigma[F_3^*](\mathbf{r}_n + \sigma \Phi_1 \bar{\mathbf{x}}_n)$
       $= \dfrac{\beta_2(\mathbf{r}_n + \sigma \Phi_1 \bar{\mathbf{x}}_n)}{\max(\beta_2 \mathbf{1}_I, |\mathbf{r}_n + \sigma \Phi_1 \bar{\mathbf{x}}_n|)}$
8:      $\mathbf{x}_{temp} = \mathbf{x}_n - \tau(D^T \mathbf{p}_{n+1} + \Phi_0^T \mathbf{q}_{n+1} + \Phi_1^T \mathbf{r}_{n+1})$
9:      $\mathbf{x}_{n+1} = prox_\tau[G](\mathbf{x}_{temp})$
       $= sign(\mathbf{x}_{temp}) \max(|\mathbf{x}_n - \mathbf{x}_{temp}| - \tau\lambda, 0)$
10:   $\bar{\mathbf{x}}_{n+1} = \mathbf{x}_{n+1} + \theta(\mathbf{x}_{n+1} - \mathbf{x}_n)$
11:   end
---

*C. CP Algorihtm for Dictionary Update*

The CDL problem in Eq.(2) involves two alternate minimizations, namely the minimization with respect to $\{\mathbf{x}_{m,v}\}$ and the minimization with respect to dictionary filters $\{\mathbf{d}_m\}$. The former minimization involves solving the multiple measurement vector version of the unconstrained CSC problem, and can be easily addressed because the problems for each $v$ are decoupled from each other. While for $\{\mathbf{d}_m\}$ that requires solving the problem in Eq.(3), the problems for different $v$ are coupled, so the dictionary update is more challenging.

In addition, we should pay attention to the constrain of $\{\mathbf{d}_m\}$. As we described above, there is an implicit zero-padding of $\{\mathbf{d}_m\}$ when computing convolution operation using FFT in the CSC procedure, the zero-padding operator needs to be explicitly represented when solving $\{\mathbf{d}_m\}$ to ensure that the resulting filters from the optimization have the same spatial domain as the $\mathbf{x}_{m,v}$. Therefore, the constraint should not only involves the normalization constraint but also the zero-padding operation. The most straightforward way is to amend the constraint set $\Omega$ as

$$\Omega_{PN} = \left\{ \mathbf{d}_m \in \mathbb{R}^N : (I - PP^T)\mathbf{d}_m = 0, \|\mathbf{d}_m\|_2 = 1 \right\}, \quad (32)$$

where $P$ indicates the zero-padding operator and $P^T$ means truncation. For convenience, we switch the $\mathbf{x}_{m,v}$ to $\mathbf{x}_{v,m}$ and exchange the positions of two variables since the convolution operation satisfies the commutative law, and rewrite the dictionary update as follows

$$\arg\min_{\{\mathbf{d}_m\}} \frac{1}{2} \sum_{v=1}^{V} \left\| \sum_{m=1}^{M} \mathbf{x}_{v,m} * \mathbf{d}_m - \mathbf{s}_v \right\|_2^2 \quad s.t. \ \mathbf{d}_m \in \Omega_{PN} \ \forall m. \quad (33)$$

Then we transform the constrained minimization into a unconstrained form via an indicator function $\ell_{\Omega_{PN}}(\cdot)$, where $\ell_{\Omega_{PN}}$ is defined as

$$\ell_{\Omega_{PN}}(X) = \begin{cases} 0 & if \ X \in \Omega_{PN} \\ \infty & if \ X \notin \Omega_{PN} \end{cases}, \quad (34)$$

thus the unconstrained dictionary update is described as follows

$$\arg\min_{\{\mathbf{d}_m\}} \frac{1}{2} \sum_{v} \left\| \sum_{m} \mathbf{x}_{v,m} * \mathbf{d}_m - \mathbf{s}_v \right\|_2^2 + \sum_{m} \ell_{\Omega_{PN}}(\mathbf{d}_m) \quad (35)$$

Since we have seen the convenience of CP algorithm in solving CSC problems, it is natural to solve the dictionary update in Eq.(35) by using CP algorithm. If we define $X_{v,m} \mathbf{d}_m = \mathbf{x}_{v,m} * \mathbf{d}_m$, and

$$X = \begin{pmatrix} X_{0,0} & X_{0,1} & \cdots \\ X_{1,0} & X_{1,1} & \cdots \\ \vdots & \vdots & \ddots \end{pmatrix} \quad \mathbf{d} = \begin{pmatrix} \mathbf{d}_0 \\ \mathbf{d}_1 \\ \vdots \end{pmatrix} \quad \mathbf{S} = \begin{pmatrix} \mathbf{s}_0 \\ \mathbf{s}_1 \\ \vdots \end{pmatrix}, \quad (36)$$

we can rewrite the dictionary update as follows

$$\arg\min_{\mathbf{d}} \frac{1}{2} \|X\mathbf{d} - \mathbf{S}\|_2^2 + \ell_{\Omega_{PN}}(\mathbf{d}). \quad (37)$$

To derive the CP algorithm for dictionary update in Eq.(37), let

$$K = X, \mathbf{y} = X\mathbf{d}, \ F(\mathbf{y}) = \frac{1}{2}\|\mathbf{y} - \mathbf{S}\|_2^2, \ G(\mathbf{d}) = \ell_{\Omega_{PN}}(\mathbf{d}), \quad (38)$$

we can obtain the convex conjugates of $F$ and its proximal mapping as follows

$$F^*(\mathbf{q}) = \max_{\mathbf{q}'} \left\{ \langle \mathbf{q}, \mathbf{q}' \rangle - \frac{1}{2} \|\mathbf{q}' - \mathbf{S}\|_2^2 \right\} \quad (39)$$

$$= \frac{1}{2}\|\mathbf{q}\|_2^2 + \langle \mathbf{q}, \mathbf{S} \rangle$$

$$prox_\sigma[F^*](\mathbf{q}) = \arg\min_{\mathbf{q}'} \left\{ \frac{1}{2}\|\mathbf{q}'\|_2^2 + \langle \mathbf{q}', \mathbf{S} \rangle + \frac{\|\mathbf{q} - \mathbf{q}'\|_2^2}{2\sigma} \right\} \quad (40)$$

$$= \frac{\mathbf{q} - \sigma \mathbf{S}}{1 + \sigma}.$$

Applying Eq.(6), we have

$$prox_\tau[G](\mathbf{d}) = \arg\min_{\mathbf{d}'} \left\{ \ell_{\Omega_{PN}}(\mathbf{d}') + \frac{\|\mathbf{d} - \mathbf{d}'\|_2^2}{2\tau} \right\}. \quad (41)$$

It is clear from the geometry that

$$prox_\tau[G](\mathbf{d}) = \frac{PP^T \mathbf{d}}{\|PP^T \mathbf{d}\|_2}. \quad (42)$$

Substituting into the generic CP framework, we have the CP-based dictionary update and summary it in **Algorithm 4.**

| | |
|---|---|
| 1: | **Algorithm 4**  CP-based dictionary update |
| 2: | $L = \|(X)\|_2; \tau = 1/L; \sigma = 1/L; \theta = 1$ |
| 3: | Initialize $\mathbf{d}_0$, $\mathbf{q}_0$ to zero values |
| 4: | $\bar{\mathbf{d}}_0 = \mathbf{d}_0, \tilde{\mathbf{d}}_0 = \mathbf{d}_0$ |
| 5: | for $n = 1$: iterN |
| 6: | $\mathbf{q}_{n+1} = (\mathbf{q}_n + \sigma(X\mathbf{d}_n - \mathbf{S}))/(1+\sigma)$ |
| 7: | $\tilde{\mathbf{d}}_{n+1} = \mathbf{d}_n - \tau X^T \mathbf{q}_{n+1}$ |
| 8: | $\mathbf{d}_{n+1} = \dfrac{PP^T \tilde{\mathbf{d}}_{n+1}}{\left\| PP^T \tilde{\mathbf{d}}_{n+1} \right\|_2}$ |
| 9: | $\bar{\mathbf{d}}_{n+1} = \mathbf{d}_{n+1} + \theta(\mathbf{d}_{n+1} - \mathbf{d}_n)$ |
| 10: | end |

Based on the derivations of CSC above and dictionary update, we deservedly get the CP based CDL method (CDL-CP), in which both phases (CSC and dictionary update) are solved using CP algorithm. The complexity of dictionary update is of $\mathcal{O}(MV)$.

## IV. EXPERIMENTAL RESULTS

In [22], the authors proposed an effective solution for CSC problem in the ADMM framework to improved the computational efficiency, and developed the corresponding dictionary learning method. They have proved that their proposed CSC method is more efficient than SM method [19], and FISTA [13], and the proposed CDL method has better performance than CG method [19], ISM method [19], and SM-cns method in [36]. Given the previous comparison results demonstrated in the literatures [19, 22, 36], we only compare our proposed algorithms with the state-of-the-art methods described in [22], which uses the ADMM framework but improve the efficiency of SM method in convolutional fitting step. For the sake of argument, we mark the competing CSC method and the CDL method in [22] as CSC-ADMM and CDL-ADMM, respectively. Then we compare our proposed CSC-CP and CSCATV-CP with CSC-ADMM in terms of in terms of convergence speed and denoising tasks. Finally, we compare the proposed CDL-CP with CDL-ADMM. All ADMM based methods in [22] were computed using a publicly available implementation shared at GitHub repository [48]. All methods are implemented using MATLAB. All experiments were conducted on a PC equipped with an Intel(R) Core(TM) i7-8700 K 3.7 GHz CPU.

### A. CSC Results

In the CSC experiments, to evaluate the performance of our proposed CSC methods, a clear normalized $256 \times 256$ woman image and its corresponding noisy image were used. The noisy images are contaminated by white Gaussian noise with standard deviation of 0.001 (low noise level case) and 0.003 (high noise level case), respectively, as shown in Fig.1. Following the statements in previous literature [18-20, 22], CSC is more effective for high-frequency information than low-frequency information, thus we apply all CSC methods to high pass filtered image, which was obtained by subtracting a lowpass image. A simple mean filter with window size of $3 \times 3$ is utilized for the low pass image. For a fair comparison, a pretained $8 \times 8 \times 16$ dictionary was utilized in all CSC methods.

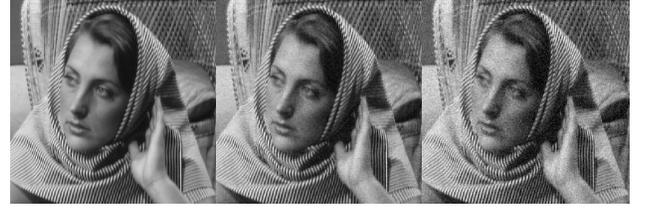

(a)Reference image  (b) noisy image with low noise level  (c) noisy image with high noise level

Fig.1 Reference and corrupted noisy images.

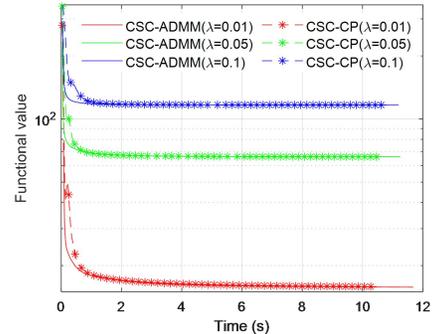

(a) Functional curve

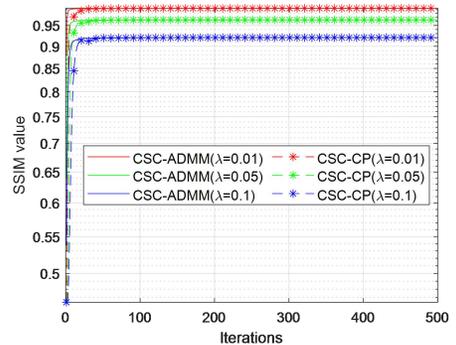

(b) SSIM curve

Fig.2 Functional values (a) and SSIM values (b) over time for the proposed CSC-CP method and the competing CSC-ADMM method using different $\lambda$ for the the noise-free image.

Fig. 2 (a) exhibits the functional values of the proposed CSC-CP and CSC-ADMM over time (500 iterations) by different $\lambda$ for the noise-free image. The other parameters in ADMM method was chosen according to [22]. Since CSC-CP and CSC-ADMM solve for the same model, here we only compare CSC-CP and CSC-ADMM, and the CSCATV-CP method is not involved. We can see that for different $\lambda$, both CSC-CP and CSC-ADMM can converge and the function value curves of CSC-CP and CSC-ADMM are almost overlapped, while the CSC-CP takes less time than the competing CSC-ADMM, although CSC-ADMM is also a single-layer iterative algorithm. This is because there are necessary element-wise multiplications on Fourier coefficients (complex) in CSC-ADMM that our approach does not require. This indicates that the proposed CSC-CP is equally effective but more efficient for all $\lambda$. Furthermore, we also calculate the SSIM values of images obtained by CSC-ADMM and CSC-CP, and plot the SSIM curves along with iterations, as shown in Fig. 2(b). As it can be seen, two SSIM curves are consistent, further demonstrating that the proposed CSC-CP has the same effectiveness with CSC-ADMM for noise-free image.

In addition, we evaluate the performances of all CSC methods on image denoising problem for two different noise levels. Fig. 3 show the PSNR values of the proposed CSC-CP, CSCATV-CP ($\beta_1=\beta_2=0.00001$), and CSC-ADMM over iterations (500 iterations) by different $\lambda$ for two noisy images. As we can see from Fig.3(a), in the case of low noise level, each method can achieve a stable PSNR value, and the PSNR curves from CSC-CP and CSC-ADMM are overlapped, while the PSNR curve from CSCATV-CP is a little higher than those of CSC-CP and CSC-ADMM. In the case of high noise level, It is obvious that CSCATV-CP obtains the highest PSNR. We can conclude from Fig.3 that CSCATV-CP performs best, and CSC-CP is comparable with CSC-ADMM in denoising performance, for all this, CSC-CP has less computational expense as what we can see from Fig.2.

($\beta_1=\beta_2$), are all selected by trial and error to achieve the maximum PSNR. From Fig.4 we can observe that all methods can obtain good denoising results in the case of low noise level, and the results are visually similar. However, in the case of high noise level, there are obvious additional artifacts in the result by CSC-ADMM even it can reduce a lot of noise. In contrast, the proposed CSC-CP and CSCATV-CP can provide better images without introducing so many additional artifacts. We also assess the results quantitatively using the PSNR and SSIM indices, as illustrated in Table 1. We can see that for the low noise level case, the maximum PSNR value that can be achieved by each method is nearly equal, and also for SSIM index, while in the case of high noise level, PSNR and SSIM values obtained by CSC-CP and CSCATV-CP are higher than those of CSC-ADMM, and CSCATV-CP obtains the highest ones. The comparisons in Fig.4 and Table 1 indicate that the proposed CSC methods provide comparative performances with CSC-ADMM in the case of low noise level, but better performance in terms of visual effects and evaluation indices for the high noise level.

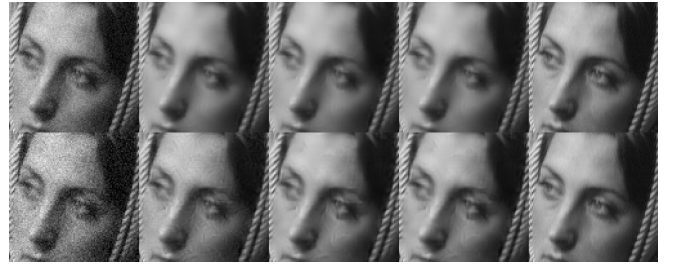

Fig.4 Denoising results by different methods under two noise levels. The top row and the bottom row respectively show the results under different low noise levels. From left to right, noisy images, results from CSC-ADMM, CSC-CP, CSCATV-CP, and reference image, respectively.

TABLE I
PSNR AND SSIM RESULTS OF THE NOISY AND PROCESSED IMAGES IN FIG. 4

|  |  | CSC-ADMM | CSC-CP | CSCATV-CP |
|---|---|---|---|---|
| low noise level | PSNR | 32.7666 | 32.7646 | **32.7669** |
|  | SSIM | 0.9191 | 0.9191 | **0.9194** |
| high noise level | PSNR | 29.8808 | 29.9381 | **30.0649** |
|  | SSIM | 0.8975 | 0.9134 | **0.9144** |

### B. CDL Results

The CDL experiments were performed using a training dataset of 40 gray images with size of $256 \times 256$. We compare the functional values of proposed CDL-CP and CDL-ADMM in terms of different dictionary size $M$ and training dataset size $V$. The complexity of dictionary update in the CDL-ADMM method is also of $\mathcal{O}(MV)$, same as that in CDL-CP. However, as what we analyze in CSC update phase above, the CP-based dictionary update in CDL-DP is also improved compared with that in CDL-ADMM. Fig. 5 shows the functional values over time for 150 iterations with different $M$ for fixed $V = 5$. All initial dictionaries with different $M$ are constructed using Gaussian random method, and the filter size is 8×8. As it can be seen, CDL-CP takes less time than CDL-ADMM, especially obvious when the $M$ is large.

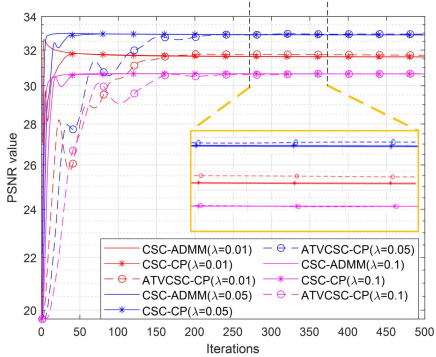

(a) PSNR curves for noisy image with low noise level

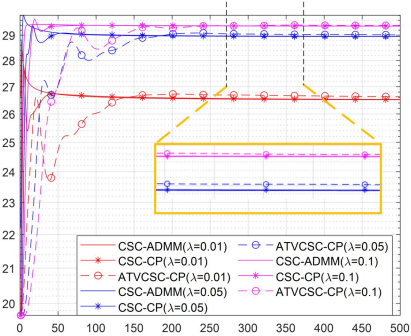

(b) PSNR curves for noisy image with high noise level

Fig.3 PSNR values with iterations for the proposed CSC-CP, CSCATV-CP, and the competing CSC-ADMM method using different $\lambda$ for the noisy images.

To clearly observe the denoisng effective of different methods, we exhibit the zoomed ROI (region of interesting) results of different methods in Fig.4. The hyperparameter parameters, including $\lambda$, iterN in each method, penalty parameter $\rho$ in CSC-ADMM, $\beta_1$ and $\beta_2$ in CSCATV-CP

Fig. 6 displays the functional values over time (150 iterations) with different *V* for fixed *M* = 16, 32, 64, 128. We can observe that given a fixed *M*, CDL-CP method is substantially faster and especially evident when more images are used to be trained. We also can see that when fewer images are trained (*V*=5), the functional values of CDL-CP achieve equivalent values with those of CDL-ADMM, but a little higher when more images are trained. Fig.7(a) and (b) shows the trained dictionaries by CDL-ADMM and CDL-CP with 150 iterations in the case of *M* = 64 and *V*=10. We can see that the structures in (a) are clearer than those in (b), but when we increase iterations, we can obtain a clear dicitonary as CDL-ADMM, as shown in Fig.7 (c). That's to say, although CDL-CP has higher computational efficiency in one iteration, but it takes more iterations to get a good dicitonary.

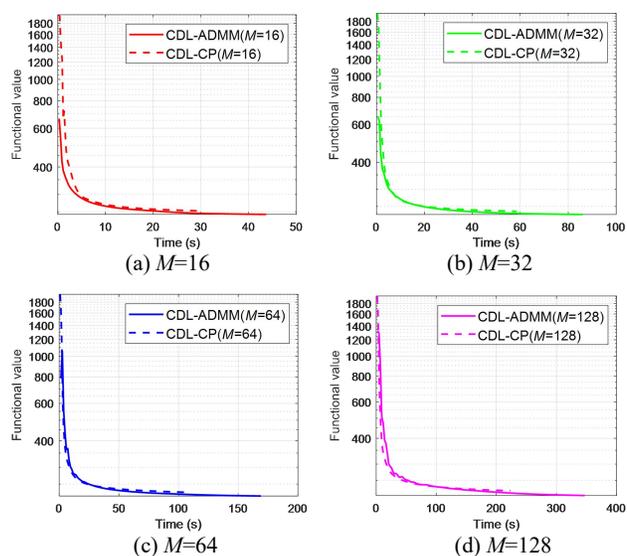

Fig.5 A comparison of functional values with time using different values of *M*, *V*= 5, filter size of 8 × 8, $\lambda = 0.05$ .

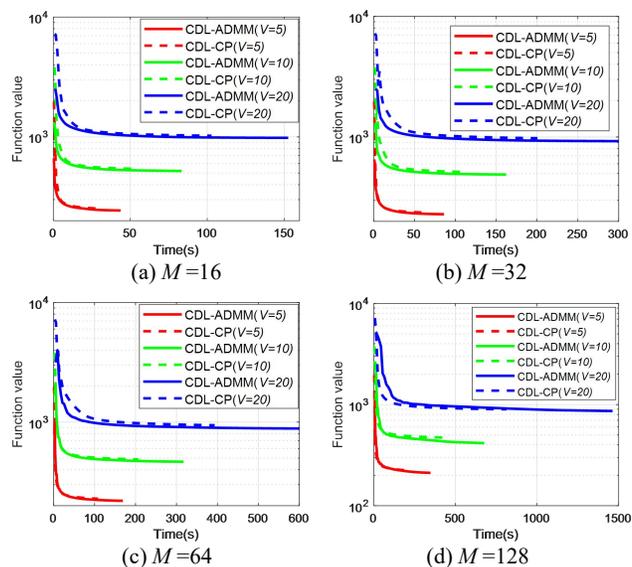

Fig.6 A comparison of functional values with time using different values of *V* for fixed *M* = 16, 32, 64, 128, filter size of 8 × 8, $\lambda = 0.05$ .

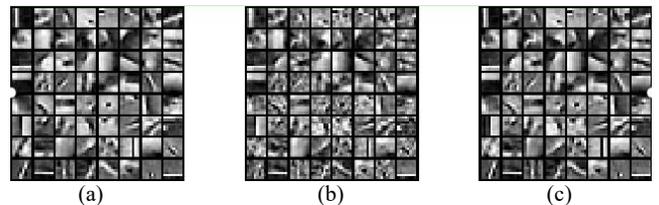

Fig.7 Dictionary comparisons: (a) CDL-ADMM with 150 iterations, (b) CDL-CP with 150 iterations, and (c) CDL-CP with 800 iterations. *V*=10, *M* = 64, filter size of 8 × 8, $\lambda = 0.05$ .

## V. CONCLUSION

In this paper, we propose a novel efficient CSC method (CSC-CP) for minimizing Eq. (1) by introducing CP framework, which breaks the conventional practice of using ADMM framework to solve CSC problem. Using the CP framework, the procedure of CSC solving only involves a single-layer loop by using the proximal mapping, and can improve the computational efficiency. What's more, parameters ( $L, \tau, \sigma$ ) in the proposed CSC-CP can be automatically determined, what we need to set up carefully is only parameter $\lambda$ in Eq. (1). Whereas in the ADMM based CSC solving processing, there is an very important parameter $\rho$ need to be carefully set. Although the CSC-ADMM method has a a scheme that can automatically set parameter $\rho$ , however it introduces another two hperparameters, more details can be found in [19]. Fig. 2(a) shows that the proposed CSC-CP method has advantages in computational efficiency, while Fig. 2(b) shows that the proposed CSC-CP method has the comparable effectiveness to CSC-ADMM in representing a noise-free image. For noisy images, the denoising effect of CSC-CP is similar to that of CSC-ADMM, while when the noise is larger, the denoising effect of CSC-CP is better, as shown in Fig. 4 and Table 1. Furthermore, to enhance the performance of convolutional sparse representation, we propose an anisotropic TV penalty to sparsely constrain the coefficient maps and derive the corresponding CSC solving in the CP framework. The experimental results show that CSCATV-CP method has better denoising effect than CSC-ADMM and CSC-CP method, and the effect is boosted for noisy images with high noise level, as shown in Fig. 4 and Table 1.

In the case of CDL, we also present a new way to update the convolutional dicctionary using CP algorithm (**Algorithm 4**), and during the dictionary training we use the proposed CSC-CP to update the convolutional coffeicient maps. Fig. 5 and Fig. 6 show that the proposed CDL-CP method leads to substantially reduced computational cost. However, the CDL-CP method in slightly inferior to the CDL-ADMM method by conducting the same iterations, as can been seen from the functional curves in Figs.5-7, and we need more iterations to train a considerable dictionary as that of CDL-ADMM, as shown in Fig.7(c). Even so, this paper provides a new strategy to solve CDL probelm at a much smaller computation time penalty in one iteration.

This manuscript presents novel efficient convolutional sparse coding algorithms and convolutional dictionary learning algorithm, which can be further used in image processing tasks, such as image denoisng, reconstruction, fusion and so on.